\documentclass[runningheads,a4paper]{llncs}

\usepackage{amsmath}
\usepackage{amssymb}
\usepackage{xparse}
\usepackage{xspace}
\usepackage{url}
\usepackage{graphicx}
\usepackage{subfig}
\usepackage{mathtools}
\usepackage{booktabs}
\usepackage{xcolor}
\usepackage{float}
\usepackage{mciteplus}

\def\andothers{et al.\,}
\def\figname{Fig.\,}

\usepackage{tikz}
\usepackage{multicol}
\usetikzlibrary{shapes,arrows}

\tikzstyle{data} = [rectangle, draw, fill=blue!20, 
    text width=5em, text centered, rounded corners, minimum height=4em, node distance=2.5cm]
\tikzstyle{line} = [draw, -latex']
\tikzstyle{proc} = [draw, rectangle, fill=gray!20, text width=6em, text centered, node distance=2.5cm, minimum height=5em]
\tikzstyle{param} = [draw, ellipse, fill=gray!40, text width=4em, text centered, node distance=2.5cm, minimum height=5em]
\tikzstyle{optparam} = [draw, ellipse, dashed, fill=gray!40, text width=4em, text centered, node distance=2.5cm, minimum height=5em]
\tikzstyle{optproc} = [draw, rectangle, dashed, fill=gray!20, text width=6em, text centered, node distance=2.5cm, minimum height=5em]
\title{Gesture Recognition in RGB Videos Using Human Body Keypoints and Dynamic Time Warping}

\graphicspath{{images/}}

\author{Pascal Schneider, Raphael Memmesheimer, Ivanna Kramer and Dietrich Paulus%
}
\institute{Active Vision Group, Institute for Computational Visualistics,
        University of Koblenz-Landau, 56070 Koblenz, Germany \\
{\{pschneider, raphael, ivannamyckhal, paulus\}@uni-koblenz.de},\\ %home page:
\texttt{http://homer.uni-koblenz.de}, \texttt{http://agas.uni-koblenz.de}}

\begin{document}

\maketitle
\thispagestyle{empty}
\pagestyle{empty}

\begin{abstract}
    Gesture recognition opens up new ways for humans to intuitively interact with machines. Especially for service robots, gestures can be a valuable addition to the means of communication to, for example, draw the robot's attention to someone or something. Extracting a gesture from video data and classifying it is a challenging task and a variety of approaches have been proposed throughout the years.
    This paper presents a method for gesture recognition in RGB videos using \emph{OpenPose} to extract the pose of a person and \emph{Dynamic Time Warping} (DTW) in conjunction with \emph{One-Nearest-Neighbor} (1NN) for time-series classification. The main features of this approach are the independence of any specific hardware and high flexibility, because new gestures can be added to the classifier by adding only a few examples of it. We utilize the robustness of the Deep Learning-based OpenPose framework while avoiding the data-intensive task of training a neural network ourselves. We demonstrate the classification performance of our method using a public dataset.
\end{abstract}

\section{Introduction}
\label{sec:introduction}
% \begin{figure}[t!]
%   \begin{multicols}{2}
%     \includegraphics[width=\linewidth]{images/sample24_crop.jpg}\par 
%     \includegraphics[width=\linewidth]{images/sample24Skeleton_crop.jpg}\par 
%   \end{multicols}

% \label{fig:exampleSkeleton}
% \end{figure}

\begin{figure}[t] \centering$
  \vspace{0.06in}
  \begin{array}{cc}
      \includegraphics[height=.45\linewidth]{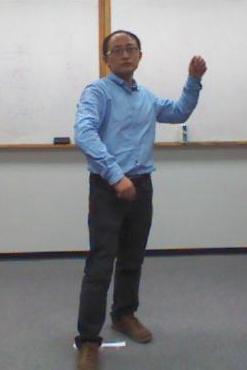} &
      \includegraphics[height=.45\linewidth]{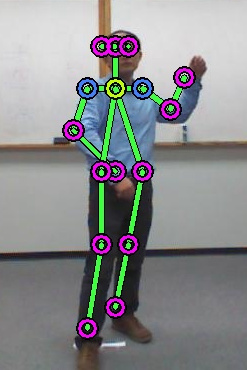}
  \end{array}$
  \caption{Example for an extracted pose using OpenPose from the UTD-MHAD dataset \cite{chen2015utd}.}
  \label{fig:exampleSkeleton}
\end{figure}

Gesture recognition is an active field of research with applications such as automatic recognition of sign language, interaction of humans and robots or for new ways of controlling video games. The main application we have in mind is an accessible way to use gestures for interacting with service robots.

Deep learning-based approaches have set new records in classification tasks in terms of their performance throughout the last few years. Consequently, they have also been applied to the problem of gesture recognition, where they could also provide good results. However, this usually comes at the cost of being very data-intensive. As with many deep learning techniques, good performance can usually only be reached with large amounts of labeled training samples. Our goal is therefore to present an approach which allows adding new gestures to the classifier with minimal effort. The training process we employ significantly reduces the overhead. Moreover, removing a gesture from the model does not involve any further cost, whereas for many other machine learning algorithms this would entail re-training the entire model \cite{gillian2011recognition}.

Moreover, we want to avoid the need of any specific hardware. For example, the \emph{Microsoft Kinect} is a popular platform for training models and collecting data for gesture recognition \cite{rosa2016fast} \cite{jiang2015multi} \cite{ribo2016approach} \cite{bautista2012probability}, since it provides not only an RGB video but also depth data. The task of gesture recognition can be simplified by placing special markers on the person's body \cite{mitra2007gesture} or special gloves for hand gestures \cite{kevin2004trajectory}. Since the main application we have in mind is human-robot interaction for service robots, relying on installing hardware on humans or manipulating the environment beforehand are impractical.

We present a method that completely avoids proprietary platforms and the need of specific hardware and instead only relies on RGB video that can be recorded using any camera with reasonable video quality in attempt to make gesture recognition more accessible. The key idea is to combine the capability of the deep learning-based OpenPose framework for extracting poses from color images and DTW, a well-established method for time-series classification.

% \usepackage{todonotes}
% \todo{Contribution}

The paper is structured as follows: in Section \ref{sec:related_work} we give an overview of some recently proposed methods for gesture recognition as well as a selection of relevant papers for both OpenPose and Dynamic Time Warping. In Section \ref{sec:approach} we describe our approach, which is summarized in \figname \ref{fig:pipeline}. In Section \ref{sec:experiments} we present the results of our experiments. We conclude our findings in Section \ref{sec:conclusion} and motivate possible future research in Section \ref{sec:futurework}.

\section{Related Work}
\label{sec:related_work}
Using Dynamic Time Warping for gesture recognition is an established approach \cite{reyes2011featureweighting}\cite{ribo2016approach}\cite{ten2007multi}. For time-series classification (TSC) in general, DTW in combination with a \emph{One-Nearest-Neighbor} (1NN) classifier has shown to provide very strong performance \cite{xi2006fast}\cite{bagnall2017great}.
DTW has been prominently used in the field of speech recognition since the 1970's. A lot of research has been focused on reducing the computational complexity, e.g. by introducing global constraints such as the \emph{Sakoe-Chiba-Band} \cite{sakoe1978dynamic}, the \emph{Itakura-Parallelogram} \cite{itakura1975minimum} or the \emph{Ratanamahatana-Koegh-Band} \cite{ratanamahatana2004making}. Notable work in improving the performance of DTW has also been done by Salvador and Chan, who proposed an approximation of DTW with linear time and space complexity \cite{salvador2007toward}. 

A detailed description of Dynamic Time Warping and the constraints is beyond the scope of this paper and hence omitted here. Introductions to the DTW algorithm and some extensions can be found in \cite{muller2007information} and \cite{senin2008dynamic}.

The growing popularity of deep learning has also influenced research in the field of gesture recognition. The method presented in this paper only uses deep learning for extracting the pose of people, not for the classification. Others have presented neural network architectures to address the problem of gesture recognition directly as a whole. Examples include the \emph{Two Streams Recurrent Neural Network} proposed by Chai \andothers \cite{chai2016two} or the \emph{Recurrent 3D Convolutional Neural Network} (R3DCNN) by Molchanov \andothers \cite{molchanov2016online}.

The problem of recognizing gestures in a video or any other sequence of data can be split into two sub-problems: segmentation and recognition. A sequence of data might contain any number of gestures, therefore the individual gestures have to be segmented first. If both segmentation and recognition are performed, it is commonly referred to as \emph{continuous} gesture recognition. Whereas if only recognition is done, this is called \emph{isolated} gesture recognition. Our method only addresses the latter. An approach to extending a DTW-based gesture recognition to the continuous case is given in \cite{reyes2011featureweighting}.

A general survey on different gesture recognition techniques can be found in \cite{liu2018gesture}, also including DTW as an approach.

There has also been research on performing gesture recognition on single RGB images using \emph{Convolutional Pose Machines} and different supervised learning techniques \cite{memmesheimer2018gesture}. The authors concluded that an extension towards using sequences rather than single images could presumably lead to significant improvements. 
% \edino{ $\Leftarrow$ self-citing, do we want this?}

Our work is focused on how the human poses extracted by OpenPose can be processed and used as input signals for Dynamic Time Warping in such a way that these two components form a processing pipeline which ultimately yields a classification of human gestures using only RGB images. What sets this apart from proposed methods based on the \emph{Microsoft Kinect} \cite{celebi2013gesture}\cite{rosa2016fast}\cite{ribo2016approach} is that we do not use depth data and extract the pose key points ourselves instead of relying on ones provided by the Kinect framework. This makes our approach independent of any special sensor hardware.

Rwigema \andothers proposed an approach to optimize weights for gesture recognition when using weighted DTW \cite{rwigema2019differential}. They also used the \emph{UTD-MHAD} dataset to verify the performance of their method and achieved an accuracy of 99.40\%. The key difference compared to our method is their choice of data to perform the recognition on. While we restrict ourselves to only the RGB video, Rwigema \andothers aimed at a multi-sensor setup using skeleton joint frames and data from a depth sensor and inertial sensor.
% section related_work (end)

\section{Approach}
\label{sec:approach}

\begin{figure} \centering
  \begin{tikzpicture}[node distance = 2cm, auto]
    % Place nodes
    \node [data] (rgbseq) {Sequence of RGB images};
    \node [proc, below of=rgbseq] (openpose) {OpenPose (with COCO body model)};
    \node [data, below of=openpose] (keypoints) {Sequence of key points};
    \node [proc, below of=keypoints] (normalization) {Key point normalization};
    \node [data, below of=normalization] (normkeypoints) {Normalized key points};
    \node [proc, below of=normkeypoints] (dimsel) {Smoothing and dimension selection};
    \node [proc, below of=dimsel] (dtw) {Dynamic Time Warping};
    \node [data, below of=dtw] (warpdist) {Warping distances};
    \node [proc, below of=warpdist] (1nn) {Nearest neighbor classifier};

    \node [data, left of=keypoints] (shoulderdist) {Distance between shoulder key points};
    \node [data, right of=keypoints] (neckpos) {Position of neck key point};
    \node [data, right of=normkeypoints] (exgest) {Example gesture key points};

    \node [param, left of=normkeypoints] (varthresh) {Variance threshold};
    \node [param, left of=rgbseq] (opparams) {Network resolution};
    \node [optparam, left of=warpdist] (classthresh) {Classifi-cation threshold};

    \node [optproc, right of=normalization] (extrain) {Train examples};

    % Draw edges
    \path [line] (rgbseq) -- (openpose);
    \path [line] (openpose) -- (keypoints);
    \path [line] (keypoints) -- (normalization);
    \path [line] (normalization) -- (normkeypoints);
    \path [line] (normkeypoints) -- (dimsel);
    \path [line] (dimsel) -- (dtw);
    \path [line] (dtw) -- (warpdist);
    \path [line] (keypoints) -- (shoulderdist);
    \path [line] (keypoints) -- (neckpos);
    \path [line] (shoulderdist) -- (normalization);
    \path [line] (neckpos) -- (normalization);
    \path [line] (exgest) -- (dimsel);
    \path [line] (varthresh) -- (dimsel);
    \path [line] (warpdist) -- (1nn);
    \path [line] (opparams) -- (openpose);
    \path [line] (classthresh) -- (1nn);
    \path [line] (extrain) -- (exgest);
  \end{tikzpicture}
  \caption[]{Overview of the processing pipeline of our method. (Grey rectangles represent processing steps, blue rectangles represent data, ellipses represent parameters.)} \label{fig:pipeline}
\end{figure}
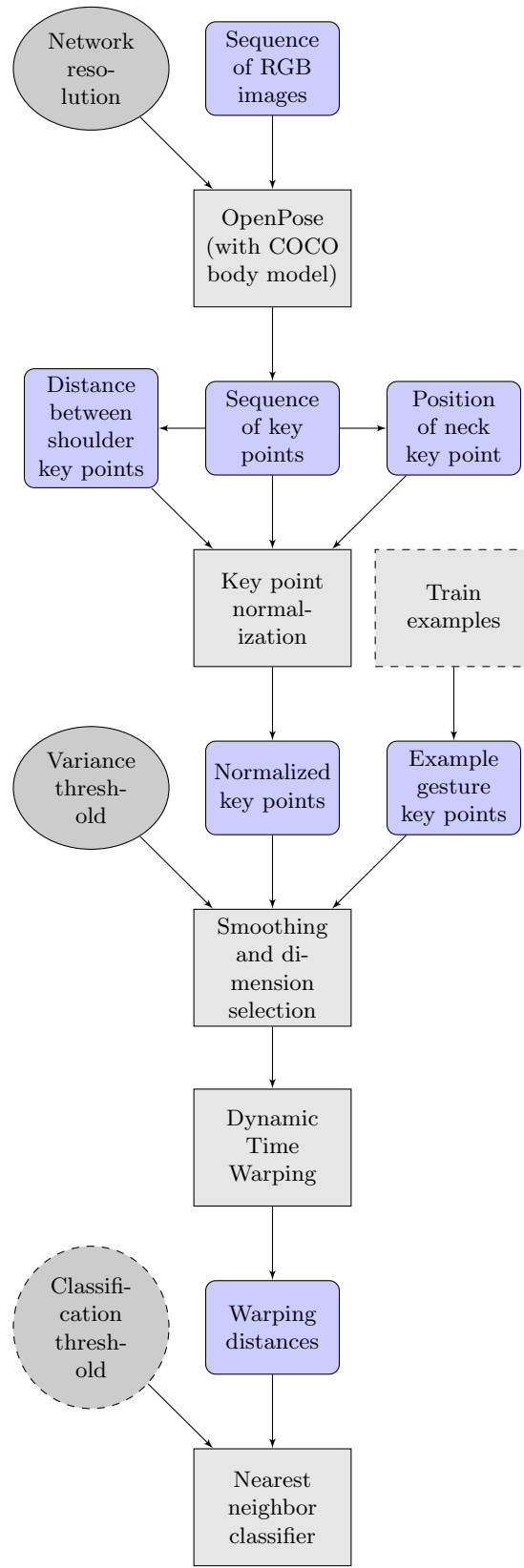

\figname \ref{fig:pipeline} shows the basic processing pipeline of the proposed method approach. Its individual steps will be detailed in the following.

\subsection{Recording RGB Videos}
Avoiding the need for special hardware is one of the key aspects of the method we want to present. We therefore only use RGB videos. The image quality and resolution have to be sufficiently high to enable OpenPose to reliably extract the pose key points. Moreover, the video frame rate has to be high enough to provide adequate spatial resolution of the signals. Most customary web cams will nowadays meet this requirement which we hope will make this method very accessible.

\subsection{Pose Estimation}
To extract the pose, we use the OpenPose framework, which is based on Convolutional Pose Machines \cite{wei2016cpm}. It features different pose models such as \emph{MPI}, \emph{COCO} and \emph{BODY\_25}. We chose the \emph{COCO} model, because we consider its 18 key points to provide a good trade-off between a detailed representation of the human pose and complexity. OpenPose also supports extracting key points for the face, hands \cite{simon2017hand} and feet \cite{cao2018openpose}, but for our application aimed at full-body gestures these key points add hardly any useful information while greatly increasing the computational complexity.

\subsection{Normalization}
The pose key points from OpenPose are given in image coordinates. We normalize the key points first before passing them on to the DTW classifier to achieve scale invariance and translational invariance. This is necessary, because otherwise the key points' coordinates are dependent on the position of the person was standing relative to the camera.
We ignore rotational invariance, since we consider this to be much less relevant, because humans can be expected to be in a mostly upright position under normal circumstances. However, adding rotational invariance might be necessary if tilt of the camera has to be corrected for.

The normalization is a simple coordinate transformation done in two steps:
\begin{enumerate}
  \item \textbf{Translation:} All the key points are translated such that the neck key point becomes the origin of the coordinate system. This is achieved by subtracting the neck key points coordinates from all other key points.
  \item \textbf{Scaling:} The key points are scaled such that the distance between the left shoulder and right shoulder key point becomes $1$. This is done by dividing all key points coordinates by the distance between the left and right shoulder key point.
\end{enumerate}

The scale normalization is inspired by Celebi \andothers \cite{celebi2013gesture}. It can be easily seen that this way of normalizing the scale can fail when the person is not oriented frontal to the camera since the shoulder-to-shoulder distance we consider here is not the actual distance in the world but instead its 2D projection onto the image plane. This leads to an important assumption of our approach: the person performing the gesture has to be oriented (roughly) frontal to the camera.

\figname \ref{fig:exampleSkeleton} shows an example of an extracted pose skeleton for a video frame from the \emph{UTD-MHAD} dataset \cite{chen2015utd}. The neck and shoulder key points are highlighted due to their importance for the normalization.

\subsection{Train Examples} \label{sec:selectExample}
We employ a simple One-Nearest-Neighbor classifier, which has proven to work well with Dynamic Time Warping in the context of time-series classification \cite{xi2006fast}\cite{bagnall2017great}. A classification that relies on comparing directly against a set of labeled examples does not need a training stage per se. New gestures could be added simply by adding an example of it. Yet it can be beneficial to incorporate a training step to find the best examples to include for the classifier. Such a method is described by Gillian \andothers \cite{gillian2011recognition}. A number of examples for the same gesture is recorded and the examples are compared to each other using the same DTW algorithm used by the classifier. The example with the minimum total warping distance to all other examples for the same gesture is chosen. This can be thought of as choosing the example which represents the gesture the best.

Instead of selecting a single example from the recorded ones for each gesture, you could also use all of them and switch to using a \emph{k-nearest-neighbor} classifier instead of 1NN. The key argument against this approach is that the computational complexity of the classification grows linearly with the number of examples each sequence has to be compared to. Therefore, we try to limit the number of examples where possible.

\subsection{Smoothing and Dimension Selection}
For most gestures, only parts of the body are relevant. Hence, only a few of the key points might be relevant for each gesture. Take for example a \emph{wave-with-left-hand} gesture: only the key points of the left arm are of relevance here and the others will usually be uninformative.
This observation can be used to reduce the dimensionality of the problem at this step. If all key point sequences were to be used in the DTW, this would total up to 36 dimensions (18 key points with an x- and y-coordinate each). The neck key point coordinates will always be uninformative due to the normalization. To further reduce the number of signals to be processed by DTW, we perform a dimension selection. This step is greatly inspired by the work of ten Holt \andothers

The criterion to select a dimension is the variance of its signal. Key points that do not move significantly during a gesture will cause the signals of the respective coordinates to be roughly constant with only little variance. All signals whose variance is below a threshold will be filtered out and are assumed to be uninformative. This filtering is done for the sequence to be classified as well as for the example sequences of each gestures. The set of dimensions for which DTW algorithm is then performed is the union of those dimensions for which the variance is above the threshold for either the sequence to be classified or the example sequence. If only those dimensions were considered where the variance is above the threshold for the sequence to be classified, some combinations of gestures could pose problems. Consider for example if there were a \emph{wave-with-left-hand}, \emph{wave-with-right-hand} and a \emph{wave-with-both-hands} gesture. Classify a newly recorded \emph{wave-with-left-hand} correctly is problematic if only its salient dimensions would be used in the DTW.
Variance in the signal might also be due to noise. A noticeable source of noise was observed caused by the limited spatial resolution of the output from OpenPose. The quantization error caused sudden spikes in the signal. We therefore smooth the signal first before determining whether it should be included for the DTW. We use a median filter with radius $r=3$ for the smoothing. The decision whether a dimension will be included for the DTW is done on the median filtered signal. However, the signal used for further processing is instead filtered using a Gaussian filter with $\sigma = 1$. This is done because the median filter is very effective at removing the noise spikes, but edges in the resulting signal are overly brought out. This worsened classification performance in our experiments, while the Gaussian filter is able to mitigate noise without these adverse effects.

In a last step before the DTW, the mean of the signal is subtracted from it, thus making it \emph{zero-mean}. A common step for feature scaling is to also normalize the signal to have \emph{unit-variance} by dividing the signal by its standard deviation. However, this had an adverse effect on classification performance in our experiments, possibly because differences in the amplitude of key point coordinate signals is relevant for classification. We therefore only transform the signals to to zero-mean, but \emph{not} to unit-variance.

\subsection{Dynamic Time Warping}
We employ the \emph{FastDTW} method by Salvador and Chan \cite{salvador2007toward} to perform DTW on each selected dimension separately. Their method is aimed at providing an approximation of DTW with less computational cost compared to the classical DTW algorithm. Finding the optimal warping path is not guaranteed with this method, but we consider this limitation to be outweighed by the superior computational performance. For the internal distance metric FastDTW uses we chose Euclidean distance. The result is the warping distances of the sequence to be classified to all gesture examples of the classifier. 

\subsection{Classification}
The classification is done using a simple \emph{One-Nearest-Neighbor Classifier} (1NN). The metric used for determining the nearest neighbor is the warping distance. A new sequence is classified to a gesture class by calculating the warping distance to all training examples and choosing the class of the training sample for which the warping distance is minimal. An additional threshold can be used in order not to classify a gesture sequence to any class if it does not resemble any of the example gesture sequences. If the minimal warping distance is still very high, this sequence can be considered to contain none of the known gestures.

\section{Experiments}
\label{sec:experiments}
\subsection{Dataset}
To evaluate the performance of our method we chose the multi-modal human action dataset of the University of Texas at Dallas (\emph{UTD-MHAD} \cite{chen2015utd}). Each gesture is performed by eight subjects four times each. Since we want to operate on RGB data, we use the color videos they provide. These videos feature a resolution of 640x480 pixels at around 30 frames per second. 
Due to the limitations of the normalization method, we specifically selected gestures where the shoulder-to-shoulder key point distance remains roughly constant throughout the sequence. The selected gestures are given by Table \ \ref{tab:actionsFromMHAD}.

\begin{table}[t!]
    \begin{center}
        \begin{tabular}{ | c | l |}
            \hline
            \textbf{Identifier} & \textbf{Action description}        \\ \hline
            a1                  & right arm swipe to the left        \\ \hline
            a6                  & cross arms in the chest            \\ \hline
            a7                  & basketball shoot                   \\ \hline
            a9                  & right hand draw circle (clockwise) \\ \hline
            a24                 & sit to stand                       \\ \hline
            a26                 & forward lunge (left foot forward)  \\ \hline
        \end{tabular}
        \caption{Selected actions from the \emph{UTD-MHAD} \cite{chen2015utd} dataset to perform classification on}
        \label{tab:actionsFromMHAD}
    \end{center}
\end{table}

\subsection{Key Point Signals}
\begin{figure}[t!]
    \centering{
    \includegraphics[width=.7\linewidth]{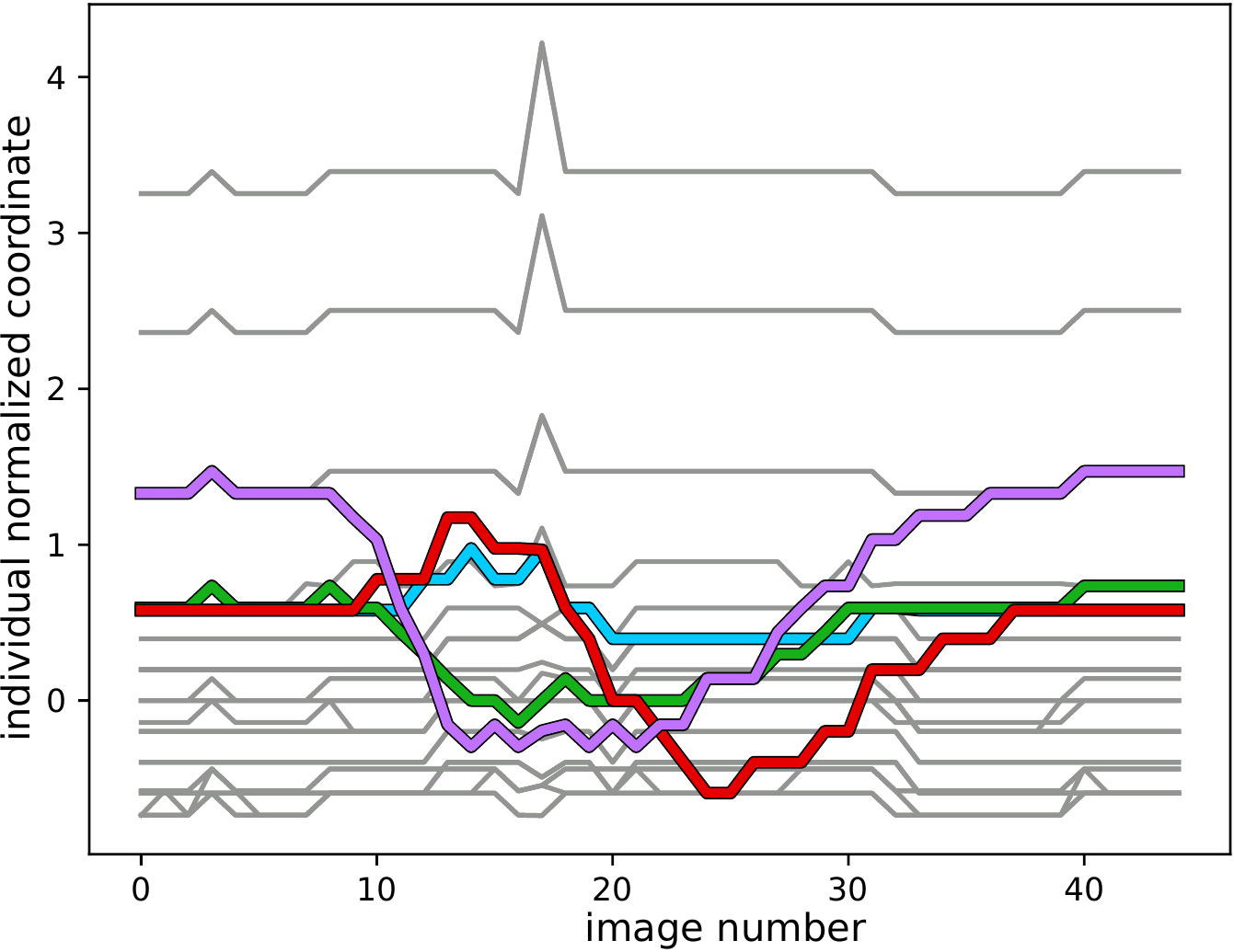}
    }
    \caption{Normalized key point coordinates for a sequence of 44 images from a person performing the \emph{right arm swipe to the left} gesture in the \emph{UTD-MHAD} dataset. The salient dimensions are highlighted.}
    \label{fig:keypointsignals}
\end{figure}

\figname \ref{fig:keypointsignals} shows the signals for every normalized coordinate of the extracted pose for a video sequence consisting of 44 images, i.e. \ it shows a separate signal for each x- and y-coordinate of each key point. Since the COCO body model has 18 key points, there are 36 individual signals. The video sequence shows a person performing the \emph{right arm swipe to the left} gesture. Most signals are roughly constant throughout the sequence. However, four of the dimensions are highlighted in \figname \ref{fig:keypointsignals}, since they can be considered salient and provide especially good signal shapes for DTW to work with. Unsurprisingly, these dimensions belong to the x- and y-coordinate of the left hand and left arm key point.

The variations in the signals for key points of body parts which are not being moved during the \emph{right arm swipe to the left} gesture (such as legs etc.) are mostly due to the noise caused by the limited resolution of the extracted pose. A conspicuous noise spike can be seen at frame 17. It is caused by the neck key point being located at a slightly higher position for one frame. Since the neck key point is the origin of our normalized coordinate system, it has a noticeable effect across multiple dimensions. A median filter with radius $r=3$ will filter out most of these spikes, which is the reason why we introduce this filtering step.

% \begin{figure}[t!]
%  \centering{
%     \includegraphics[width=.7\linewidth]{images/confMat_130-of-168_wo_a8.pdf}
%     }
%     \caption{Confusion matrix for the classification of the actions given in Table\ \ref{tab:actionsFromMHAD}. Actual classes are on the horizontal axis, predicted classes on the vertical axis.}
%     \label{fig:CM_without_a8}
% \end{figure}

% \begin{figure}[t!]
% \centering{
%     \includegraphics[width=.7\linewidth]{images/confMat_125-of-196_w_a8.pdf}
%     }
%     \caption{Confusion matrix for the classification when action \emph{a8} is also added.}
%     \label{fig:CM_with_a8}
% \end{figure}

\begin{figure}[t] \centering$
  \vspace{0.06in}
  \begin{array}{cc}
      \includegraphics[width=.5\linewidth]{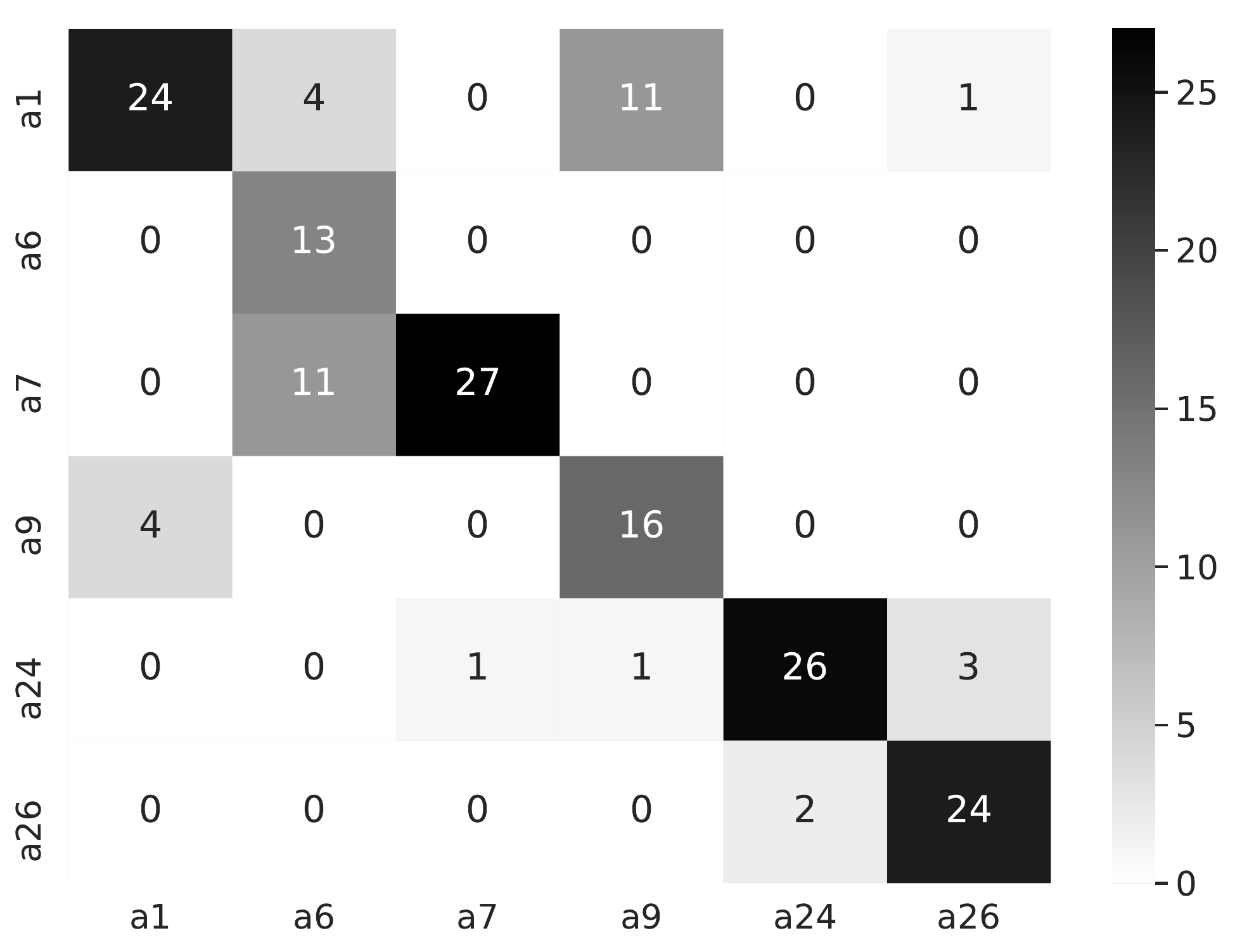} &
      \includegraphics[width=.5\linewidth]{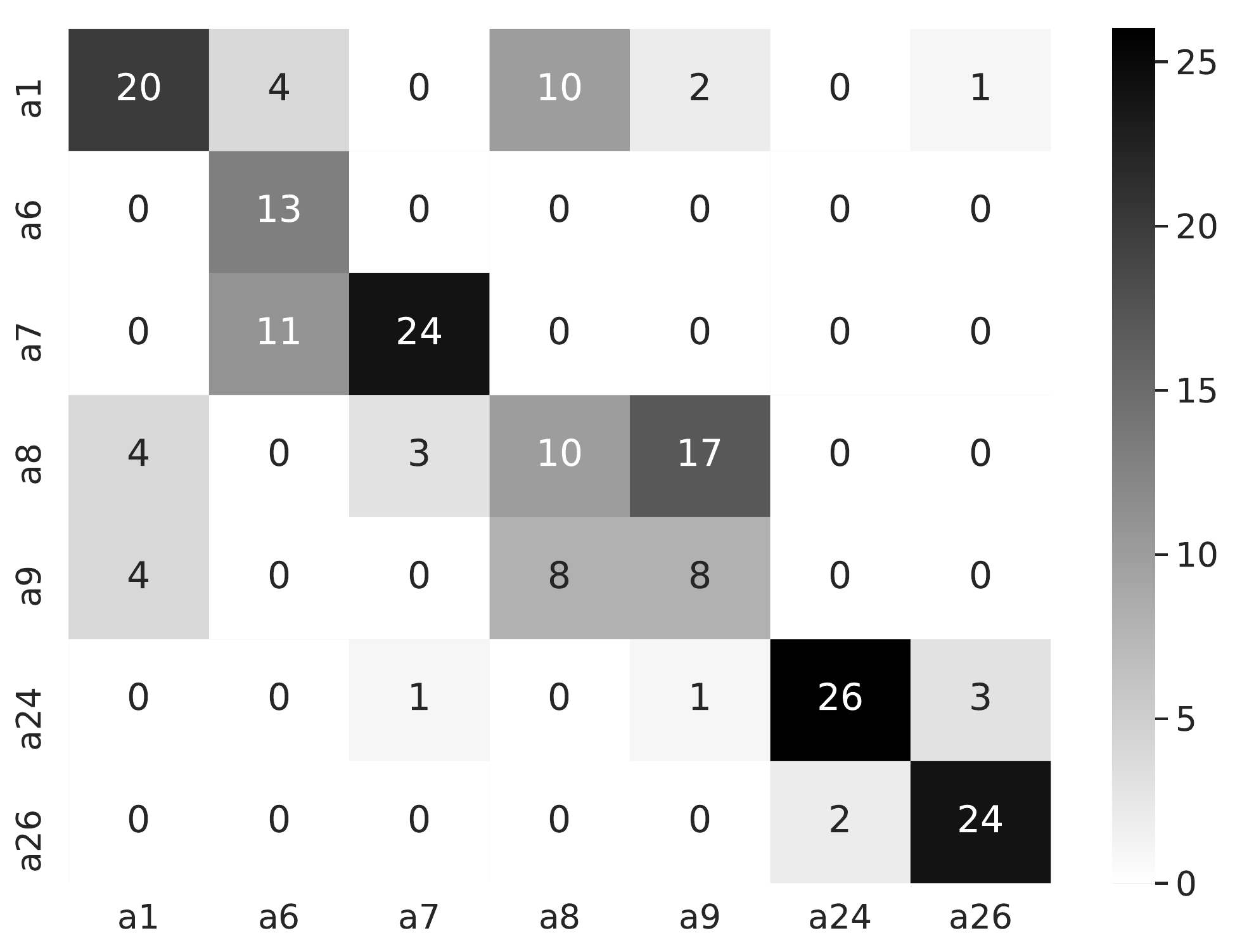}\\
      (a) & (b) 
  \end{array}$
  \caption{Confusion matrix for the classification of the actions given in Table\ \ref{tab:actionsFromMHAD}. Actual classes are on the horizontal axis, predicted classes on the vertical axis $(a)$. Confusion matrix for the classification for performance of \emph{a8} $(b)$.}
  \label{fig:CM}
\end{figure}

\subsection{Classification Performance}
We selected six different gestures from the \emph{UTD-MHAD} dataset. The selected gestures are given in Table \ \ref{tab:actionsFromMHAD}. To select an example for each gesture, we only considered the gesture performances by subject one, i.e.\ from the four sequences for subject one for each of the gestures, one is selected as described in Section \ref{sec:selectExample}. The other three sequences are not considered for the classification. From the 168 sequences which were classified, 130 were classified correctly. This equates to approximately 77.4\%. The confusion matrix is given by \figname \ref{fig:CM} $(a)$.

To further test the discriminative strength of the classification, we added another gesture to the classification: gesture \emph{a8, right hand draw x}. The confusion matrix for this experiment is illustrated in \figname \ref{fig:CM} $(b)$. As can be seen, the classification performance deteriorated significantly. 125 of 196 sequences were classified correctly (63.8\%). Most notably, \emph{a9} was classified as \emph{a8} more often than it was classified correctly. This clearly shows the limitations of the method. Gestures that are too similar to each other can not be distinguished.

An important parameter for the processing is the variance threshold $t_\mathrm{var}$. Choosing a very low threshold will result in many dimensions being selected for the DTW step, which makes computation slow. If the threshold is set too high on the other hand, possibly none of the signals will exceed it and the classification will fail because no data reaches it. Table \ref{tab:differentVTs} shows the classification performance for different values of $t_\mathrm{var}$. Finding the appropriate value for $t_\mathrm{var}$ is not part of the method, so it has to be chosen a priori.  We can not derive any general advice for how to choose $t_\mathrm{var}$ from this data, this question could be addressed in future research.

\begin{table}[t!]
    \begin{center}
        \begin{tabular}{| c | c | c |}
            \cline{2-3}
            \multicolumn{1}{c|}{}     & \multicolumn{2}{|c|}{\textbf{Correctly classified}}                    \\ \hline
            \textbf{$t_\mathrm{var}$} & \textbf{without a8}                                 & \textbf{with a8} \\ \hline
            0.05                      & 72.0\%                                              & 63.3\%           \\ \hline
            0.10                      & 77.4\%                                              & 63.8\%           \\ \hline
            0.15                      & 74.4\%                                              & 67.9\%           \\ \hline
            0.20                      & 76.2\%                                              & 66.8\%           \\ \hline
        \end{tabular}
        \caption{Percentage of correctly classified gestures for different variance thresholds $t_\mathrm{var}$}
        \label{tab:differentVTs}
    \end{center}
\end{table}

\section{Conclusion}
\label{sec:conclusion}
We presented a method for gesture recognition on RGB videos using OpenPose to extract pose key points and Dynamic Time Warping in conjunction with a One-Nearest-Neighbor classifier to perform classification. We showed how this can be used to perform gesture recognition with only very little training data and without the need for special hardware. Our first tests using this method yielded promising results if the gestures where sufficiently different, but also revealed limitations in case of attempting to classify more similar gestures.

Recent methods for gesture recognition using multi-modal data are often able to outperform our results in terms of accuracy, even more so considering our focus on only few selected gestures. Examples include the method by Rwigema \andothers \cite{rwigema2019differential} with an accuracy of 99.40\% on the \emph{UTD-MHAD} dataset, Celebi \andothers \cite{celebi2013gesture} with an accuracy of 96.70\% on their own dataset or Molchanov \andothers \cite{molchanov2016online} with up to 83.8\% accuracy, also using their own custom dataset. Nonetheless, we find our results promising considering the substantially reduced amount of data available to our method by restricting ourselves to only RGB videos, which is often the most easily obtainable data in a real-world scenario.

\section{Future Work}
\label{sec:futurework}
A variety of modifications to the original DTW algorithm have been presented through the years. Some have already been mentioned in Section \ref{sec:related_work}. Others include for example methods for adding feature weighting \cite{reyes2011featureweighting}, \emph{Derivative Dynamic Time Warping (D-DTW)} \cite{keogh2001derivative} or \emph{Multi-Dimensional} DTW \cite{ten2007multi}. The effect these modifications have on the performance of nearest neighbor classifiers based on warping distance is often times not obvious and also dependent on factors like noise in the signal \cite{ten2007multi}. Future research could try to work out general guidelines for when to use which variant of DTW.
In addition to these fundamental algorithmic options, there are a number of other factors which can impact the classification performance parameters, such as the window size of DTW, the variance threshold or the choice of example gestures.

Only single-person gesture recognition has been regarded in this paper. Since OpenPose is also capable of detecting the poses of multiple people at once, upgrading to multi-person gesture recognition is a possible subject for future research.
The \emph{UTD-MHAD} dataset we used for our experiments was recorded in a very controlled environment, further tests should be conducted to find out how our results generalize to more realistic scenarios.

Another topic of research is how this method can be sped up, desirably up to the point where it reaches real-time capability.

% \subsubsection*{Acknowledgement}

\bibliographystyle{IEEEtranM}

\bibliography{references}

\end{document}